\documentclass[11pt,a4paper]{article}
\usepackage[hyperref]{acl2019}
\usepackage{times}
\usepackage{latexsym}
\usepackage{graphicx}
\usepackage{amsmath}
\usepackage{booktabs}
\usepackage{multirow}
\usepackage{csquotes}

\usepackage{url}

\aclfinalcopy 

\title{Universal Text Representation from BERT: An Empirical Study}

\author{Xiaofei Ma, Zhiguo Wang, Patrick Ng, Ramesh Nallapati, Bing Xiang \\
AWS AI Labs \\
 {\tt \{xiaofeim, zhiguow, patricng, rnallapa, bxiang\}@amazon.com}  \\}

\date{}

\begin{document}
\maketitle
\begin{abstract}
We present a systematic investigation of layer-wise BERT activations for general-purpose text representations to understand what linguistic information they capture and how transferable they are across different tasks.
Sentence-level embeddings are evaluated against two state-of-the-art models on downstream and probing tasks from SentEval, 
while passage-level embeddings are evaluated on four question-answering (QA) datasets under a learning-to-rank problem setting.
Embeddings from the pre-trained BERT model perform poorly in semantic similarity and sentence surface information probing tasks.
Fine-tuning BERT on natural language inference data greatly improves the quality of the embeddings.
Combining embeddings from different BERT layers can further boost performance.
BERT embeddings outperform BM25 baseline significantly on factoid QA datasets at the passage level,
but fail to perform better than BM25 on non-factoid datasets.
For all QA datasets, there is a gap between embedding-based method and in-domain fine-tuned BERT (we report new state-of-the-art results on two datasets),
which suggests deep interactions between question and answer pairs are critical for those hard tasks.
\end{abstract}

\section{Introduction}
Universal text representations are important for many NLP tasks as modern deep learning models are becoming more and more data-hungry and computationally expensive.
On one hand, most research and industry tasks face data sparsity problem due to the high cost of annotation.  
Universal text representations can mitigate this problem to a certain extent by performing implicit transfer learning among tasks.
On the other hand, modern deep learning models with millions of parameters are expensive to train and host,
while models using text representation as the building blocks can achieve similar performance with much fewer tunable parameters.
The pre-computed text embeddings can also help decrease model latency dramatically at inference time. 
 
Since the introduction of pre-trained word embeddings such as word2vec \cite{Mikolov} and GloVe \cite{Pennington2014},
a lot of efforts have been devoted to developing universal sentence embeddings.
Initial attempts at learning sentence representation using unsupervised approaches did not yield satisfactory performance.
Recent work \cite{Conneau2017} has shown that models trained in supervised fashion on datasets like Stanford Natural Language Inference (SNLI) corpus \cite{Bowman2015a}
can consistently outperform unsupervised methods like SkipThought vectors \cite{Kiros2015}.
More recently, Universal Sentence Encoder \cite{Cer2018} equipped with the Transformer \cite{Vaswani2017} as the encoder,
co-trained on a large amount of unsupervised training data and SNLI corpus,
has demonstrated surprisingly good performance with minimal amounts of supervised training data for a transfer task.
  
BERT \cite{Devlin2018}, one of the latest models that leverage heavily on language model pre-training,
has achieved state-of-the-art performance in many natural language understanding tasks ranging from sequence and sequence pair classification to question answering.
The fact that pre-trained BERT can be easily fine-tuned with just one additional output layer to create a state-of-the-art model for a wide range of tasks
suggests that BERT representations are potential universal text embeddings.

Passages that consist of multiple sentences are coherent units of natural languages that convey information at a pragmatic or discourse level.
While there are many models for generating and evaluating sentence embeddings, 
there hasn't been a lot of work on passage level embedding generation and evaluation.

In this paper, we conducted an empirical study of layer-wise activations of BERT as general-purpose text embeddings.
We want to understand to what extent does the BERT representation capture syntactic and semantic information. 
The sentence-level embeddings are evaluated on downstream and probing tasks using the SentEval toolkit \cite{Conneau2018},
while the passage-level encodings are evaluated on four passage-level QA datasets (both factoid and non-factoid) under a learning-to-rank setting.
Different methods of combining query embeddings with passage-level answer embeddings are examined. 

\section{BERT Sentence Embedding}
We use the SentEval toolkit to evaluate the quality of sentence representations from BERT activations.
The evaluation encompasses a variety of downstream and probing tasks.
Downstream tasks include text classification, 
natural language inference, 
paraphrase detection, 
and semantic similarity. 
Probing tasks use single sentence embedding as input, 
are designed to probe sentence-level linguistic phenomena, 
from superficial properties of sentences to syntactic information to semantic acceptability.
For details about the tasks, 
please refer to  \cite{Conneau2018} and \cite{Conneau2018a}.
We compare the BERT embeddings against two state-of-the-art sentence embeddings,
Universal Sentence Encoder \cite{Cer2018},  InferSent \cite{Conneau2017},
and a baseline of averaging GloVe word embeddings. 


\textbf{Effect of Encoder Layer}:
We compare the performance of embeddings extracted from different encoder layers of a pre-trained BERT using bert-as-service \cite{xiao2018bertservice}.
Since we are interested in the linguistic information encoded in the embeddings,
we only add a logistic regression layer on top of the embeddings for each classification task.
The results of using [CLS] token activations as embeddings are presented in Figure \ref{fig:cls_pooling}.
The raw values are provided in the Appendix.
In the heatmap, the raw values of metrics are normalized by the best performance of a particular task from all the models we evaluated including BERT.
The tasks in the figure are grouped by task category.  
For example, all semantic similarity related tasks are placed at the top of the figure.

\begin{figure}[tbp]
	\includegraphics[width=0.55\textwidth]{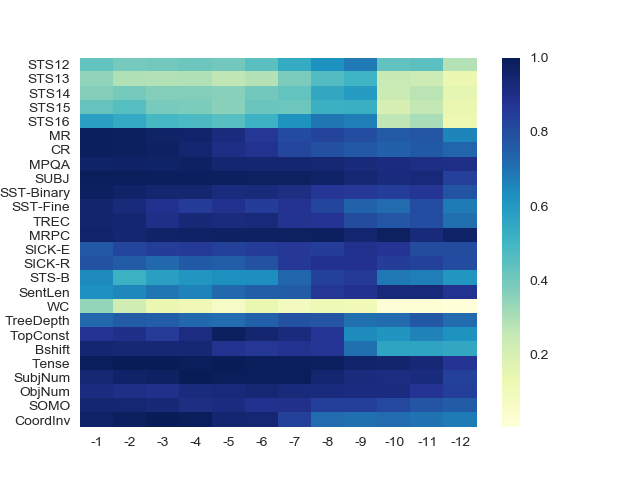}
	\caption{Sentence embedding performance of [CLS] token activation from different layers of BERT.  Color value of 1 corresponds to the best performance on a given task. Numbers on the x-axis represent the pooling layer with -1 being the top encoder layer, -12 being the bottom layer.}
	\label{fig:cls_pooling}
\end{figure}  

As can be seen from the figure, 
embeddings from top layers generally perform better than lower layers.
However, for certain semantic probing tasks such as tense classification, subject, and object number classifications,
middle layer embeddings perform the best.
Intuitively, embeddings from top layer should be more biased towards the target of BERT pre-training tasks,
while bottom layer embeddings should be close to the word embeddings.
We observed a higher correlation in performance between bottom layer embeddings and GloVe embeddings than embeddings from other layers. 
Overall, pre-trained BERT embeddings perform well in text classification and syntactic probing tasks.
The biggest limitation lies in the semantic similarity and sentence surface information probing tasks,
where we observed a big gap between BERT and other state-of-the-art models.

\textbf{Effect of Pooling Methods}:
We examined different methods of extracting BERT hidden state activations.
The pooling methods we evaluated include:
CLS-pooling (the hidden state corresponding to the [CLS] token),
SEP-pooling (the hidden state corresponding to the [SEP] token),
Mean-pooling (the average of the hidden state of the encoding layer on the time axis),
and Max-pooling (the maximum of the hidden state of the encoding layer on the time axis).
To eliminate the layer-wise effects, we averaged the performance of each pooling method over different layers.
The results are summarized in Table \ref{tab:pooling},
where the score for each task category is calculated by averaging the normalized values for the tasks within each category.
Although the activations of [CLS] token hidden states are often used in fine-tuning BERT for classification tasks, 
Mean-pooling of hidden states performs the best in all task categories among all the pooling methods.

\begin{table}[tbp]
\scalebox{0.8}{
\begin{tabular}{lcccc}
\bottomrule
\textbf{Task} & \textbf{[CLS]} & \textbf{Mean} & \textbf{Max}  & \textbf{[SEP]} \\ \hline
Semantic Similarity & 34.1 & \textbf{84.5} & 80.7 & 13.0 \\
Text Classification & 90.7 & \textbf{95.4} & 89.7 & 88.9 \\
Entailment & 72.4 & \textbf{89.3} & 87.1 & 66.1 \\
Surface Information & 45.6 & \textbf{78.9} & 47.3 & 42.8 \\
Syntactic Information & 78.2 & \textbf{86.0} & 75.7 & 72.2 \\
Semantic Information & 90.3 & \textbf{93.7} & 89.5 & 86.7 \\
\midrule
Average Score & 68.6 & \textbf{88.0} & 78.3 & 61.6 \\
\bottomrule
\end{tabular}
}
\caption{Comparison of pooling methods}
\label{tab:pooling}
\end{table}

\begin{table*}[tbp]
\scalebox{0.8}{
\begin{tabular}{lccccccccc}
\toprule
\textbf{Task} & \textbf{PT (t)} & \textbf{MNLI (t)} & \textbf{SNLI (t)}  & \textbf{PT (t+b)} & \textbf{MNLI (t+b)} & \textbf{SNLI (t+b)} & \textbf{GloVe} & \textbf{USE} & \textbf{InferSent} \\ \hline
Semantic Similarity & 82.2 & 89.8 & 97.6 & 90.6 & 94.9 & 98.5 & 76.7 & \textbf{99.1} & 95.6 \\
Text Classification  & 97.1 & 97.2 & 97.7 & 97.1 & \textbf{98.0} &  \textbf{98.0} & 92.8 & 97.5 & 95.3 \\
Entailment  & 88.8 & 92.0 & 97.5 & 92.7 & 95.2 & 98.2 & 88.2 & 97.8 & \textbf{99.2} \\
Surface Information  & 66.3 & 61.2 & 61.1	& 87.4 & \textbf{89.4} & \textbf{89.4} & 72.1 & 54.1 & 58.5 \\
Syntactic Information  & 89.5 & 85.5 & 85.9 & \textbf{94.1} & 90.6 & 92.0 & 71.0 & 71.5 & 77.6 \\
Semantic Information  & 97.0 & 96.3 & 95.9 & \textbf{97.4} & 96.4 & 96.0 & 84.6 & 88.2 & 90.7 \\
\midrule
Average Score & 86.8 & 87.0 & 89.3 & 93.2 & 94.1 & \textbf{95.4} & 80.9 & 84.7 & 86.2 \\
\bottomrule
\end{tabular}
}
\caption{Comparison across models. PT stands for pre-trained BERT. MNLI and SNLI stand for BERT fine-tuned on MNLI, SNLI, representatively. Letters in parentheses represent BERT pooling layers. \enquote{t} means top layer, \enquote{b} means bottom layer. Mean-pooling is used for all BERT embeddings. Logistic regression layer is added on top of the embeddings.}
\label{tab:pooling2}
\end{table*}

\textbf{Pre-trained vs. Fine-tuned BERT}:
All the models we considered in this paper benefit from supervised training on natural language inference datasets.
In this section, we compare the performance of embeddings from pre-trained BERT and fine-tuned BERT. 
Two natural language inference datasets, MNLI \cite{Williams2017} and SNLI, were considered in the experiment.
Inspired by the fact that embeddings from different layers excel in different tasks,
we also conducted experiments by concatenating embeddings from multiple layers.
The results are presented in Table \ref{tab:pooling2},
and the raw values are provided in the Appendix. 

As we can see from the table, 
embeddings from pre-trained BERT are good at capturing sentence-level syntactic information and semantic information,
but poor at semantic similarity tasks and surface information tasks.
Our findings are consistent with \citep{Goldberg2019} work on assessing BERT's syntactic abilities. 
Fine-tuning on natural language inference datasets improves the quality of sentence embedding,
especially on semantic similarity tasks and entailment tasks.
Combining embeddings from two layers can further boost the performance on sentence surface and syntactic information probing tasks.
Experiments were also conducted by combining embeddings from multiple layers.
However, there is no significant and consistent improvement over pooling just from two layers.
Adding multi-layer perceptron (MLP) instead of logistic regression layer on top of the embeddings also provides no significant changes in performance,
which suggests that most linguistic properties can be extracted with just a linear readout of the embeddings.
Our best model is the combination of embeddings from the top and bottom layer of the BERT fine-tuned on SNLI dataset.

\section{BERT Passage Embedding}
In this section, we evaluate BERT embeddings at passage level on question-answering datasets under a learning-to-rank problem setting.

\textbf{Datasets}:
We experimented on four datasets:
(1) WikiPassageQA \cite{Cohen2018},
(2) InsuranceQA (version 1.0) \cite{Feng2016},
(3) Quasar-t \cite{Dhingra2017},
and (4) SearchQA \cite{Dunn2017}.
They cover both factoid and non-factoid QA and different average passage length.
The statistics of the four datasets are provided in the Appendix.
To generate passage-level question-answering data from Quasart-t and SearchQA,  
we used the retrieved passages for each question from OpenQA\footnote{https://github.com/thunlp/OpenQA},
and generated question-passage relevance label based on whether the ground truth answer is contained in the passage. 

\begin{table}[h]
\begin{center}
\scalebox{0.69}{
\begin{tabular}{lccc}
\toprule
\textbf{Dataset / Model} & \multicolumn{3}{c}{\textbf{Metrics}} \\ \hline
\textbf{WikiPassageQA} & \textbf{MAP} & \textbf{P@5} & \textbf{P@10} \\ \hline
BM25 & 53.7 &19.5 & 11.5 \\
Memory-CNN-LSTM \cite{Cohen2018} & 56.1 & 20.8 & 12.3 \\
Pre-trained BERT Embedding & 55.0 & 21.6 & 13.7 \\ 
SNLI Fine-tuned BERT Embedding & 52.5 & 20.6 & 12.8 \\
In-domain Fine-tuned BERT & \textbf{74.9} & \textbf{27.2} & \textbf{15.2} \\ \hline
\textbf{InsuranceQA} & \textbf{P@1} & \textbf{P@5} & \textbf{P@10} \\ \hline
BM25 & 60.2 & 19.5 & 10.9 \\
SUBMULT+NN \cite{Wang2016g} & 75.6 & - & - \\
DSSM \cite{Huang2013a} & 30.3 & - & - \\
Pre-trained BERT Embedding & 44.9 & 17.6 & 10.6 \\
SNLI Fine-tuned BERT Embedding & 48.0 & 18.5 & 11.0 \\
In-domain Fine-tuned BERT & \textbf{78.3} & \textbf{25.4} & \textbf{13.7} \\ \hline
\textbf{Quasar-t} & \textbf{R@1} & \textbf{R@5} & \textbf{R@10} \\
BM25 & 38.7 & 59.2 & 66.0 \\
Pre-trained BERT Embedding & 48.6 & 66.6 & 71.7 \\
SNLI Fine-tuned BERT Embedding & 49.3 & 66.1 & 71.0 \\ 
In-domain Fine-tuned BERT & \textbf{59.5} & \textbf{70.9} & \textbf{74.6} \\ \hline
\textbf{SearchQA} & \textbf{R@1} & \textbf{R@5} & \textbf{R@10} \\
BM25 & 50.5 & 83.3 & 90.9 \\
Pre-trained BERT Embedding & 66.2 & 89.7 & 95.0 \\
SNLI Fine-tuned BERT Embedding & 66.8 & 90.0 & 95.1 \\
In-domain Fine-tuned BERT & \textbf{76.3} & \textbf{93.0} & \textbf{96.7} \\
\bottomrule
\end{tabular}
}
\end{center}
\caption{Results of BERT passage-level embeddings on question-answering datasets}
\label{tab:passage}
\end{table}

\textbf{Experiment Setting}:
We use the same pooling methods as in the sentence embedding experiment to extract passage embeddings,
and make sure that the passage length is within BERT's maximum sequence length.
Different methods of combining query embeddings with answer passage embeddings were explored including:
cosine similarity (no trainable parameter), bilinear function, concatenation, and $(u, v, u * v, |u - v|)$ where $u$ and $v$ are query embedding and answer embedding, respectively. 
A logistic regression layer or an MLP layer is added on top of the embeddings to output a ranking score.
We apply the pairwise rank hinge loss $l(q, +a, -a; \theta) = max\{0,  - S(q, +a; \theta)+S(q, -a; \theta)\}$ to every tuple of $(query, +answer, -answer)$.
Ranking metrics such as MRR (mean reciprocal rank), MAP (mean average precision), Precision@K and Recall@K are used to measure the performance.
We compared BERT passage embeddings against the baseline of BM25, other state-of-the-art models,
and a fine-tuned BERT on in-domain supervised data which serves as the upper bound.
For in-domain BERT fine-tuning, we feed the hidden state of the [CLS] token from the top layer into a two-layer MLP which outputs a relevance score between the question and candidate answer passage.
We fine-tune all BERT parameters except the word embedding layers.

\textbf{Results}:
The comparison between BERT embeddings and other models is presented in Table \ref{tab:passage}.
Overall, in-domain fine-tuned BERT delivers the best performance.
We report new state-of-the-art results on WikiPassageQA ($33\%$ improvement in MAP) and InsuranceQA (version 1.0) ($3.6\%$ improvement in P@1) by supervised fine-tuning BERT using pairwise rank hinge loss.
When evaluated on non-factoid QA datasets, there is a big gap between BERT embeddings and the fully fine-tuned BERT,
which suggests that deep interactions between questions and answers are critical to the task.
However, the gap is much smaller for factoid QA datasets.
Since non-factoid QA depends more on content matching rather than vocabulary matching, 
the results are kind of expected.
Similar to BERT for sentence embeddings, mean-pooling and combining the top and bottom layer embeddings lead to better performance, 
and  $(u, v, u * v, |u - v|)$ shows the strongest results among other interaction schemes.
Different from sentence-level embeddings, 
fine-tuning BERT on SNLI doesn't lead to significant improvement,
which suggests possible domain mismatch between SNLI and the QA datasets.
MLP layer usually provided a 1-2 percent boost in performance compared to the logistic regression layer. 
For WikiPassageQA, BERT embeddings perform comparably as BM25 baseline.
For InsuranceQA, BERT embeddings outperform a strong representation-based matching model DSSM \cite{Huang2013a},
but still far behind the state-of-the-art interaction-based model SUBMULT+NN \cite{Wang2016g} and fully fine-tuned BERT. 
On factoid datasets (Quasar-t and SearchQA), BERT embeddings outperform BM25 baseline significantly.

\section{Conclusion}
In this paper, we conducted an empirical investigation of BERT activations as universal text embeddings.
We show that sentence embeddings from BERT perform strongly on SentEval tasks,
and combining embeddings from the top and bottom layers of BERT fine-tuned on SNLI provides the best performance.
At passage-level, we evaluated BERT embeddings on four QA datasets.
Models based on BERT passage embeddings outperform BM25 baseline significantly on factoid QA datasets but fail to perform better than BM25 on non-factoid datasets.
We observed a big gap between embedding-based models and in-domain the fully fine-tuned BERT on QA datasets.
Future research is needed to better model the interactions between pairs of text embeddings.

\bibliographystyle{acl_natbib}
\bibliography{acl2019}

\begin{thebibliography}{19}
\expandafter\ifx\csname natexlab\endcsname\relax\def\natexlab#1{#1}\fi

\bibitem[{Bowman et~al.(2015)Bowman, Angeli, Potts, and Manning}]{Bowman2015a}
Samuel~R. Bowman, Gabor Angeli, Christopher Potts, and Christopher~D. Manning.
  2015.
\newblock \href {https://doi.org/10.18653/v1/D16-1264} {{A large annotated
  corpus for learning natural language inference}}.

\bibitem[{Cer et~al.(2018)Cer, Yang, Kong, Hua, and Limtiaco}]{Cer2018}
Daniel Cer, Yinfei Yang, Sheng-yi Kong, Nan Hua, and Nicole Limtiaco. 2018.
\newblock \href {https://doi.org/arXiv:1803.11175v2} {{Universal Sentence
  Encoder}}.
\newblock \emph{arXiv:1803.11175 [cs]}.

\bibitem[{Cohen et~al.(2018)Cohen, Yang, and Croft}]{Cohen2018}
Daniel Cohen, Liu Yang, and W~Bruce Croft. 2018.
\newblock \href {http://arxiv.org/abs/arXiv:1805.03797v1} {{WikiPassageQA: A
  Benchmark Collection for Research on Non-factoid Answer Passage Retrieval}}.
\newblock In \emph{SIGIR}, pages 1--4.

\bibitem[{Conneau and Kiela(2018)}]{Conneau2018}
Alexis Conneau and Douwe Kiela. 2018.
\newblock \href {http://arxiv.org/abs/1803.05449} {{SentEval: An Evaluation
  Toolkit for Universal Sentence Representations}}.

\bibitem[{Conneau et~al.(2017)Conneau, Kiela, Schwenk, Barrault, and
  Bordes}]{Conneau2017}
Alexis Conneau, Douwe Kiela, Holger Schwenk, Loic Barrault, and Antoine Bordes.
  2017.
\newblock \href {https://doi.org/10.1.1.156.2685} {{Supervised Learning of
  Universal Sentence Representations from Natural Language Inference Data}}.
\newblock pages 670--680.

\bibitem[{Conneau et~al.(2018)Conneau, Kruszewski, Lample, Barrault, and
  Baroni}]{Conneau2018a}
Alexis Conneau, German Kruszewski, Guillaume Lample, Lo{\"{i}}c Barrault, and
  Marco Baroni. 2018.
\newblock \href {http://arxiv.org/abs/1805.01070} {{What you can cram into a
  single vector: Probing sentence embeddings for linguistic properties}}.

\bibitem[{Devlin et~al.(2018)Devlin, Chang, Lee, and Toutanova}]{Devlin2018}
Jacob Devlin, Ming-Wei Chang, Kenton Lee, and Kristina Toutanova. 2018.
\newblock \href {http://arxiv.org/abs/1810.04805} {{BERT: Pre-training of Deep
  Bidirectional Transformers for Language Understanding}}.

\bibitem[{Dhingra et~al.(2017)Dhingra, Mazaitis, and Cohen}]{Dhingra2017}
Bhuwan Dhingra, Kathryn Mazaitis, and William~W Cohen. 2017.
\newblock \href {http://arxiv.org/abs/1707.03904} {{Quasar: Datasets for
  Question Answering by Search and Reading}}.
\newblock (2).

\bibitem[{Dunn et~al.(2017)Dunn, Sagun, Higgins, Guney, Cirik, and
  Cho}]{Dunn2017}
Matthew Dunn, Levent Sagun, Mike Higgins, V.~Ugur Guney, Volkan Cirik, and
  Kyunghyun Cho. 2017.
\newblock \href {http://arxiv.org/abs/1704.05179} {{SearchQA: A New Q{\&}A
  Dataset Augmented with Context from a Search Engine}}.

\bibitem[{Feng et~al.(2016)Feng, Xiang, Glass, Wang, and Zhou}]{Feng2016}
Minwei Feng, Bing Xiang, Michael~R. Glass, Lidan Wang, and Bowen Zhou. 2016.
\newblock \href {https://doi.org/10.1109/ASRU.2015.7404872} {{Applying deep
  learning to answer selection: A study and an open task}}.
\newblock In \emph{2015 IEEE Workshop on Automatic Speech Recognition and
  Understanding, ASRU 2015 - Proceedings}, pages 813--820.

\bibitem[{Goldberg(2019)}]{Goldberg2019}
Yoav Goldberg. 2019.
\newblock \href {http://arxiv.org/abs/1901.05287} {{Assessing BERT's Syntactic
  Abilities}}.
\newblock pages 2--5.

\bibitem[{Huang et~al.(2013)Huang, Urbana, He, Gao, Deng, Acero, and
  Heck}]{Huang2013a}
Po-sen Huang, N~Mathews~Ave Urbana, Xiaodong He, Jianfeng Gao, Li~Deng, Alex
  Acero, and Larry Heck. 2013.
\newblock \href {https://doi.org/10.1145/2505515.2505665} {{Learning Deep
  Structured Semantic Models for Web Search using Clickthrough Data}}.
\newblock \emph{the 22nd ACM international conference on Conference on
  information {\&} knowledge management}, pages 2333--2338.

\bibitem[{Kiros et~al.(2015)Kiros, Zhu, Salakhutdinov, Zemel, Torralba,
  Urtasun, and Fidler}]{Kiros2015}
Ryan Kiros, Yukun Zhu, Ruslan Salakhutdinov, Richard~S Zemel, Antonio Torralba,
  Raquel Urtasun, and Sanja Fidler. 2015.
\newblock \href {https://doi.org/10.1017/CBO9781107415324.004} {{Skip-Thought
  Vectors}}.
\newblock (786):1--11.

\bibitem[{Mikolov et~al.()Mikolov, Chen, Corrado, and Dean}]{Mikolov}
Tomas Mikolov, Kai Chen, Greg Corrado, and Jeffrey Dean.
\newblock \href {https://doi.org/10.1162/jmlr.2003.3.4-5.951} {{Distributed
  Representations of Words and Phrases and Their Compositionality}}.
\newblock pages 1--9.

\bibitem[{Pennington et~al.(2014)Pennington, Socher, and
  Manning}]{Pennington2014}
Jeffrey Pennington, Richard Socher, and Christopher Manning. 2014.
\newblock \href {https://doi.org/10.3115/v1/D14-1162} {{Glove: Global Vectors
  for Word Representation}}.
\newblock In \emph{Proceedings of the 2014 Conference on Empirical Methods in
  Natural Language Processing (EMNLP)}, pages 1532--1543.

\bibitem[{Vaswani et~al.(2017)Vaswani, Shazeer, Parmar, Uszkoreit, Jones,
  Gomez, Kaiser, and Polosukhin}]{Vaswani2017}
Ashish Vaswani, Noam Shazeer, Niki Parmar, Jakob Uszkoreit, Llion Jones,
  Aidan~N. Gomez, Lukasz Kaiser, and Illia Polosukhin. 2017.
\newblock \href {https://doi.org/10.1017/S0140525X16001837} {{Attention Is All
  You Need}}.
\newblock (Nips).

\bibitem[{Wang and Jiang(2016)}]{Wang2016g}
Shuohang Wang and Jing Jiang. 2016.
\newblock \href {http://arxiv.org/abs/1611.01747} {{A Compare-Aggregate Model
  for Matching Text Sequences}}.
\newblock (2016):1--11.

\bibitem[{Williams et~al.(2017)Williams, Nangia, and Bowman}]{Williams2017}
Adina Williams, Nikita Nangia, and Samuel~R. Bowman. 2017.
\newblock \href {http://arxiv.org/abs/1704.05426} {{A Broad-Coverage Challenge
  Corpus for Sentence Understanding through Inference}}.
\newblock pages 1112--1122.

\bibitem[{Xiao(2018)}]{xiao2018bertservice}
Han Xiao. 2018.
\newblock bert-as-service.
\newblock \url{https://github.com/hanxiao/bert-as-service}.

\end{thebibliography}
\end{document}